\documentclass[fleqn,10pt]{SelfArx} 

\definecolor{color1}{RGB}{0,0,90} 
\definecolor{color2}{RGB}{0,20,20} 

\usepackage[utf8]{inputenc} 
\usepackage{hyperref} 
\hypersetup{hidelinks,colorlinks,breaklinks=true,urlcolor=color2,citecolor=color1,linkcolor=color1,bookmarksopen=false,pdftitle={Title},pdfauthor={Author}}

\JournalInfo{Draft Version} 
\Archive{Additional note} 

\PaperTitle{Dark LLMs: The Growing Threat of Unaligned AI Models} 

\Authors{Michael Fire\textsuperscript{1}*, Yitzhak Elbazis\textsuperscript{1}, Adi Wasenstein\textsuperscript{1},  Lior Rokach\textsuperscript{1}*} 
\affiliation{\textsuperscript{1}\textit{Ben Gurion University of the Negev}} 
\affiliation{*\textbf{Corresponding authors}: mickyfi@bgu.ac.il and liorrk@bgu.ac.il} 

\newcommand{\keywordname}{Keywords} 

\Abstract{Large Language Models (LLMs) rapidly reshape modern life, advancing fields from healthcare to education and beyond. However, alongside their remarkable capabilities lies a significant threat: the susceptibility of these models to jailbreaking. The fundamental vulnerability of LLMs to jailbreak attacks stems from the very data they learn from. As long as this training data includes unfiltered, problematic, or 'dark' content, the models can inherently learn undesirable patterns or weaknesses that allow users to circumvent their intended safety controls. Our research identifies the growing threat posed by dark LLMs—models deliberately designed without ethical guardrails or modified through jailbreak techniques. In our research, we uncovered a universal jailbreak attack that effectively compromises multiple state-of-the-art models, enabling them to answer almost any question and produce harmful outputs upon request. The main idea of our attack was published online over seven months ago. However, many of the tested LLMs were still vulnerable to this attack. Despite our responsible disclosure efforts, responses from major LLM providers were often inadequate, highlighting a concerning gap in industry practices regarding AI safety. As model training becomes more accessible and cheaper, and as open-source LLMs proliferate, the risk of widespread misuse escalates. Without decisive intervention, LLMs may continue democratizing access to dangerous knowledge, posing greater risks than anticipated.}

\begin{document}

\flushbottom 

\maketitle 
\thispagestyle{empty} 


\section*{The Dual-Use Challenge} 

Large Language Models (LLMs) have rapidly become embedded in modern society, used by over a billion people,\footnote{
It is estimated that over a billion people worldwide use LLMs. OpenAI's ChatGPT has over 800 million weekly users
~\cite{digwatch2025chatgpt}, Meta's Llama models have been downloaded 650 million times~\cite{reuters2025}, and Baidu's Ernie Bot has 200 million users~\cite{wsj2025}.} accelerating discovery, democratizing knowledge, and enabling new forms of creativity. From helping researchers translate rare languages~\cite{zhang2024hire} to personalized medicine~\cite{berger2024advancing}, their positive impact is clear. However, these same models, trained on vast data, which, despite curation efforts, can still absorb dangerous knowledge, including instructions for bomb-making, money laundering, hacking, and performing insider trading~\cite{kirch2024features}. While commercial LLMs incorporate safety mechanisms to block harmful outputs, these safeguards are increasingly proving insufficient. A critical vulnerability lies in jailbreaking—a technique that uses carefully crafted prompts to bypass safety filters, enabling the model to generate restricted content~\cite{yi2024jailbreak}. 

\section*{The Rise of Dark LLMs}
Recently, a disturbing trend has gained momentum: the release of deliberately unaligned models, often described as "dark LLMs"~\cite{poireault2023dark,zvelo2024darkllms}.
Variants such as WormGPT and FraudGPT are openly advertised online for having "no ethical guardrails" and for their willingness to assist in cybercrime, fraud, and more~\cite{zvelo2024darkllms}. These models, alongside open-source systems like Llama\footnote{\url{https://www.llama.com}} and DeepSeek\footnote{\url{https://www.deepseek.com}}, can be jailbroken to remove restrictions~\cite{jiang2024wildteaming,bachwani2025deepseek,mccauley2025universal}. As model training becomes cheaper and hardware requirements diminish~\cite{appenzeller2024llmflation}, powerful LLMs may become more accessible to individuals with malicious intent. Even in the mid of 2023, there were already more than 15,800 LLMs available on platforms like Hugging Face~\cite{gao2023origin}, reflecting the rapid proliferation of these models. What was once restricted to state actors or organized crime groups may soon be in the hands of anyone with a laptop or even a mobile phone.

\section*{Jailbreaking: Unlocking Forbidden Knowledge}
Even carefully aligned LLMs are vulnerable to manipulation. Through a technique known as jailbreaking, attackers craft adversarial prompts that bypass safety filters, forcing models that cost tens of millions to create, like ChatGPT\footnote{\url{https://chatgpt.com}} and Gemini,\footnote{\url{https://gemini.google.com/app}} to output restricted content~\cite{ft2024jailbreak,andriushchenko2024jailbreaking,mccauley2025universal}. An entire ecosystem has emerged around the creation and distribution of jailbreak prompts; for instance, the ChatGPT Jailbreak subreddit alone has amassed approximately 141,000 users, referred as Jailbreakers.\footnote{\url{https://www.reddit.com/r/ChatGPTJailbreak/}} Alarmingly, recent research has demonstrated that even simple character sequences can successfully bypass safeguards in multiple leading models simultaneously~\cite{andriushchenko2024jailbreaking}. Moreover, a recent study from April 2025,  introduced a novel universal jailbreak attack capable of bypassing protections in a wide range of LLMs, including advanced reasoning models~\cite{mccauley2025universal}.  As the market for jailbreak techniques continues to expand, the potential to weaponize LLMs is no longer a theoretical risk--it is a tangible reality, easily accessible to those who seek it, even young kids and teenagers.

\section*{A Glimpse Into the Dark Potential}
Our research began by investigating the real-world implications of LLM jailbreak attacks and evaluating the defense mechanisms embedded within commercial models. We started with a publicly known jailbreak method, published over seven months ago on Reddit. Surprisingly, many of the leading LLMs we tested, including state-of-the-art commercial systems, remained vulnerable to this widely disseminated attack. We developed a more comprehensive universal jailbreak attack based on this foundational exploit. This method proved to be highly effective, successfully bypassing safety filters in nearly all the LLMs we evaluated. Once compromised, the models consistently generated responses to virtually any query, including those involving illicit and harmful activities. Disturbingly, the LLMs themselves offered examples of illegal activities spanning various domains, often accompanied by detailed, step-by-step instructions.
To responsibly disclose this vulnerability, we contacted several leading LLM providers via official channels, including bug bounty programs and direct communication. However, the response was underwhelming. Several companies did not respond at all, while others indicated that such vulnerabilities fell outside the scope of their bounty programs, suggesting we report the issue through alternative channels instead.
These findings expose a critical weakness in the current approach to LLM security: even when vulnerabilities are well-documented and actively exploited in public forums, major providers often fail to respond adequately. The ease with which these LLMs can be manipulated to produce harmful content underscores the urgent need for robust safeguards. The risk is not speculative—it is immediate, tangible, and deeply concerning, highlighting the fragile state of AI safety in the face of rapidly evolving jailbreak techniques.

\section*{The Irreversibility of Open-Source Leaks}
Unlike centrally managed platforms like ChatGPT or Gemini, open-source LLMs cannot be patched once vulnerabilities are discovered. Once an uncensored version is shared online, it is archived, copied, and distributed beyond control. No company, no update cycle, and no regulation can erase a locally saved model from a laptop or private server. Moreover, attackers can chain models together—using one model to generate jailbreak prompts for another—compounding the risk~\cite{kritz2025jailbreaking}.

\section*{What Can Be Done?}

LLM providers must actively work to patch vulnerabilities and jailbreak techniques as soon as they become known. Containing the threat of dark LLMs requires layered, proactive defenses. Key strategies include:

\begin{itemize}
    \item \textbf{Training Data Curation} - Models should be trained on curated datasets that deliberately exclude harmful content, such as bomb-making instructions, money laundering guides, and extremist manifestos. Leveraging AI-driven content screening during pretraining can significantly enhance this process. Just as we protect children from unfiltered content on TV or the internet, we should also ensure that LLMs are not exposed to dark and dangerous material.

    \item \textbf{LLM Firewalls} -  Middleware can intercept prompts and outputs, acting as a real-time safeguard between users and the model. Robust LLM firewalls should become a standard part of any deployment, just as antivirus software became ubiquitous for computers. Notably, IBM offers \textit{Granite Guardian}, a suite of models designed to detect risks in prompts and responses, ensuring safe and responsible use of large language models~\cite{padhi2024granite}. Similarly, Meta provides \textit{Llama Guard}, an open-source guardrail system aimed at building secure AI agents by detecting and mitigating harmful or inappropriate content generation~\cite{chennabasappa2025llamafirewall}.

    \item \textbf{Machine Unlearning} - Recent advances allow models to "forget" specific types of content after deployment, without full retraining~\cite{liu2025rethinking}. If perfected, machine unlearning could enable rapid removal of dangerous capabilities from already-released models.

    \item \textbf{Continuous Red Teaming} - Developers should maintain active adversarial testing teams, publish red-team performance benchmarks, and offer bug bounties for vulnerability discovery.

    \item \textbf{Public Awareness} - Governments, educators, and civil society must treat unaligned LLMs as serious security risks, comparable to unlicensed weaponry or explosives guides. Restricting casual access, especially for minors, should be a policy priority.
\end{itemize}

\section*{Conclusion: The Clock Is Ticking}
LLMs are one of the most consequential technologies of our time. Their potential for good is immense—but so is their capacity for harm if left unchecked. Unchecked, dark LLMs could democratize access to dangerous knowledge at an unprecedented scale, empowering criminals and extremists across the world. It is not enough to celebrate the promise of AI innovation. Without decisive intervention—technical, regulatory, and societal—we risk unleashing a future where the same tools that heal, teach, and inspire can just as easily destroy.
The choice remains ours. But time is running out.

\section*{Acknowledgments} 

\addcontentsline{toc}{section}{Acknowledgments} 

 While drafting this article, we used ChatGPT and Grammarly for editing.

\phantomsection
\bibliographystyle{unsrt}
\bibliography{sample}


\end{document}